\documentclass[11pt]{article}
\usepackage{acl}
\usepackage{times}
\usepackage{latexsym}
\usepackage{microtype}
\usepackage{booktabs}
\usepackage{graphicx}
\usepackage{amsmath}
\usepackage{multirow}
\usepackage{xcolor}
\usepackage{hyperref}
\hypersetup{hidelinks}
\usepackage{placeins}
\usepackage{enumitem}
\usepackage{float} 
\usepackage[most]{tcolorbox}
\definecolor{cardframe}{gray}{0.55}
\definecolor{cardback}{gray}{0.98}
\definecolor{cardtitle}{gray}{0.90}
\definecolor{ncaccent}{RGB}{178,34,34}
\tcbset{taskcardbase/.style={
    enhanced, breakable, arc=2pt, boxrule=0.3pt, leftrule=2.2pt,
    colback=cardback, colframe=cardframe,
    colbacktitle=cardtitle, coltitle=black, fonttitle=\footnotesize,
    titlerule=0pt, top=2pt, bottom=2pt, left=5pt, right=4pt, boxsep=1.5pt,
    before skip=4pt, after skip=4pt,
    lower separated=true, colbacklower=white,
    fontlower=\footnotesize\ttfamily, halign lower=flush left
}}
\newtcolorbox{taskcard}[1]{taskcardbase, title={#1}}
\newtcolorbox{taskcardnc}[1]{taskcardbase, colframe=ncaccent, colbacktitle=ncaccent!12, title={#1}}
\definecolor{promptaccent}{RGB}{47,79,110}
\newtcolorbox{promptbox}[1]{enhanced, breakable, arc=2pt, boxrule=0.3pt,
  leftrule=2.6pt, colback=promptaccent!3, colframe=promptaccent,
  colbacktitle=promptaccent!12, coltitle=black,
  fonttitle=\footnotesize\bfseries\ttfamily, titlerule=0pt,
  fontupper=\footnotesize, top=2pt, bottom=2pt, left=5pt, right=4pt,
  boxsep=1.5pt, before skip=5pt, after skip=5pt, title={#1}}
\title{{Benchmarking AI Agents for Hardware Design Automation via MCP Tool Calling}}

\author{Leonardo Liparulo \\ Politecnico di Milano \\ \texttt{leolipa02@gmail.com} \\\And Francesco Pierri \\ Politecnico di Milano \\ \texttt{francesco.pierri@polimi.it} \\}

\begin{document}
\maketitle

\begin{abstract}
We ask whether AI agents powered by locally deployed large language models can reliably automate expert-defined hardware design workflows in an industry-realistic tool-calling setting.
In these environments, engineers issue repetitive, dependency-ordered operations---such as creating components, adding ports, and wiring connections---through specialised tools. Confidentiality constraints on component specifications and naming conventions often preclude hosted proprietary APIs, motivating the use of locally deployed models.
To study this setting, we build a Model Context Protocol (MCP) server that reproduces the state and dependency logic of a proprietary hardware design tool used in embedded system development and construct a benchmark covering single-operation edits, multi-step dependency chains, invalid requests, misspelled prompts, and multi-server tool contexts.
We evaluate seven open-source models comparing pipeline choices including system prompts, tool-description detail, context scope, and single-agent versus multi-agent architectures.
Results show that strong models can achieve near-complete expected-call coverage on the benchmarked workflows, but reliability depends strongly on both task structure and agent configuration.
Comprehensive tool descriptions consistently reduce failures, few-shot prompting can cause severe inaction for some models, cumulative context harms constrained models, and multi-agent decomposition helps weak workers or long sessions at the cost of additional calls.
These findings provide practical guidance for deploying local LLM agents in stateful hardware design environments.
\end{abstract}

\section{Introduction}
\label{sec:intro}

\begin{figure}[t]
\centering
\includegraphics[width=\columnwidth]{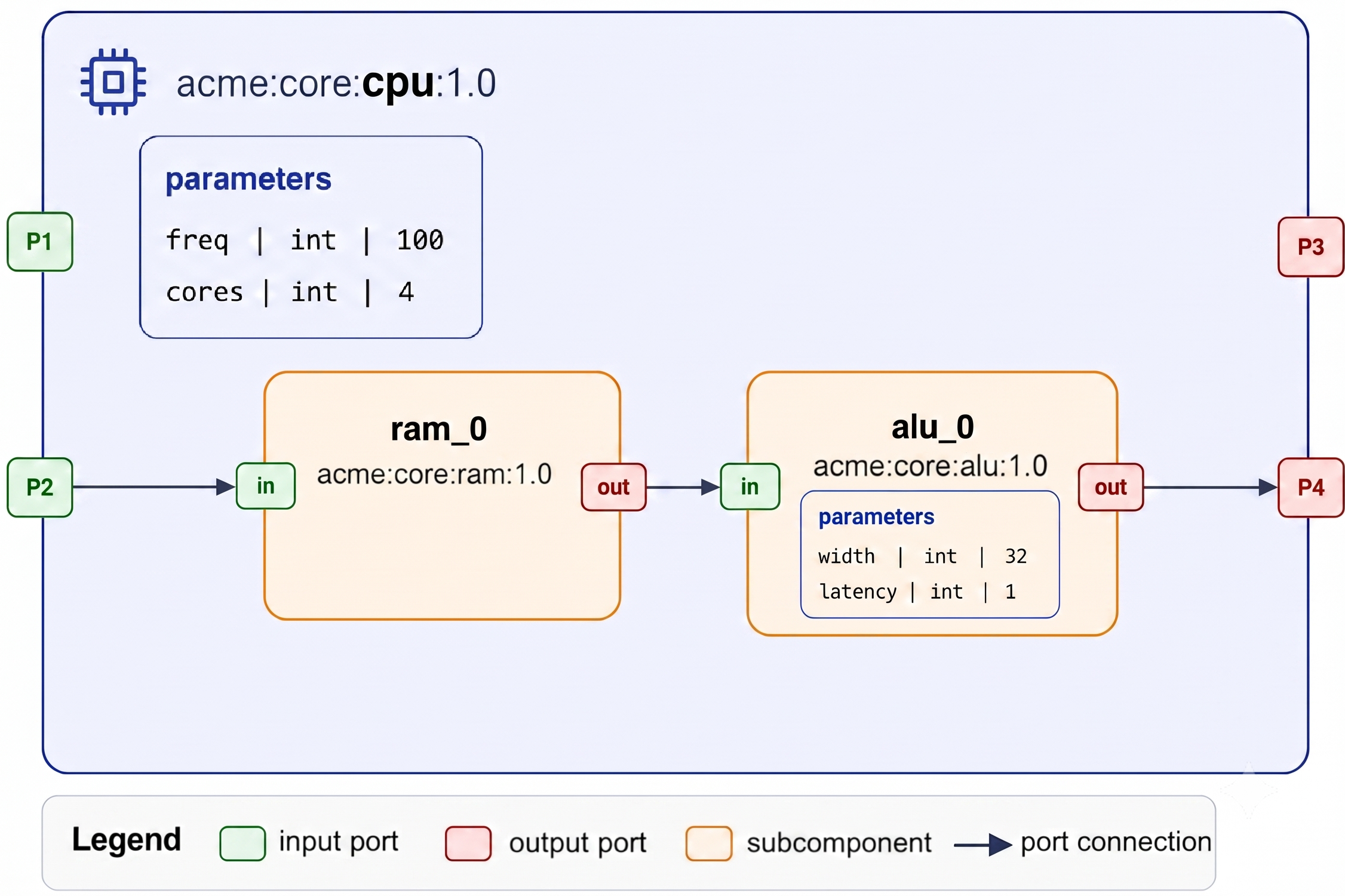}
\caption{A component in the target application domain. Subcomponents, ports, and connections form a dependency-ordered structure: each element must be created before it can be referenced by subsequent operations.}
\label{fig:component}
\end{figure}

Hardware design for embedded systems is carried out through specialised tools such as Synopsys Virtualizer or the open-source Kactus2~\cite{kactus2}, which expose graphical interfaces for building structured component models.
For instance, an engineer creates a component identified by a VLNV (Vendor, Library, Name, Version) tuple, adds signal and transactional ports, defines parameters, instantiates subcomponents, and wires connections between them.
These operations must be issued in dependency order: a port must exist before it can be connected, and a subcomponent must exist before its ports can be referenced (see Fig.~\ref{fig:component}).

In industrial practice, many of these interactions are repetitive and template-like, such as adding batches of similarly typed ports, instantiating analogous subcomponents, or wiring connections that follow a regular pattern.
Automating them is attractive because it reduces repetitive manual interaction while preserving the structured operations and dependency constraints already enforced by the design tool.

More generally, this kind of workflow consists of structured operations issued in dependency order against persistent application state.
Such settings are well-suited to LLM-based tool-calling agents, which translate natural-language requests into sequences of API calls.
Agents operating in this manner have been studied in retail and airline customer operations~\cite{yao2024tau}, CRM systems~\cite{huang2025crmarena}, enterprise workflows~\cite{drouin2024workarena}, and multi-application environments~\cite{trivedi2024appworld}.
In these settings, the agent modifies shared state through structured tool calls, and later operations depend on the results of earlier ones.
The Model Context Protocol (MCP)~\cite{mcp} standardises this style of agent--application interaction and has seen rapid adoption as an integration layer.

Hardware design shares the same stateful and dependency-ordered characteristics as these agent-driven domains, but also introduces practical deployment constraints.
Component specifications, port names, and internal naming conventions can reveal details of unreleased products, often precluding the use of hosted proprietary APIs.
Locally deployed open-source models avoid this exposure, but may be less capable than frontier-hosted models.
This raises a practical question for industrial adoption: \textbf{can LLM-based agents reliably automate expert-defined hardware design workflows in an industry-realistic tool-calling setting while operating within these constraints?}

We address this question by implementing an MCP server whose 14 tools cover the recurrent component-editing operations and dependency logic identified with professional users of a proprietary tool.
The server reproduces the relevant data model and constraints of the tool rather than calling its production APIs directly.
We then evaluate agents built on seven open-source models running locally through Ollama~\cite{ollama} as 4-bit quantised variants.

Our contributions are: (1) an MCP server for the proprietary hardware design tool;
(2) an expert-informed benchmark comprising eight task suites, spanning independent tasks, dependency chains, cross-task sessions, error handling, and multi-server contexts;
(3) a systematic evaluation across prompt, tool-description, context-management, and architecture choices; and
(4) practitioner guidelines for deploying reliable local agents in hardware design environments.

\section{Related Work}
\label{sec:related}

\noindent\textbf{Agents that operate applications.}
LLM agents that operate applications through stateful tool calls are evaluated by benchmarks such as $\tau$-bench~\cite{yao2024tau}, CRMArena~\cite{huang2025crmarena}, WorkArena~\cite{drouin2024workarena}, AppWorld~\cite{trivedi2024appworld}, and ToolSandbox~\cite{lu2025toolsandbox}.
These settings require dependency-ordered interactions over shared mutable state, structurally similar to hardware design workflows.

Because later actions depend on earlier ones, task-level success alone can hide omitted or extraneous calls, motivating call-level evaluation~\cite{yao2024tau,gao2025mcp}.
MCP-native frameworks such as MCP-RADAR~\cite{gao2025mcp} and MCP-Bench~\cite{wang2025mcp} adopt call-level scoring as a key evaluation strategy for MCP-based tool-calling systems.

LLMs have separately been applied to hardware tasks such as RTL generation~\cite{chang2023chipgpt} and EDA-flow orchestration~\cite{fu2023gpt4aigchip}, typically producing design artefacts or pipeline commands rather than operating a stateful design application through structured tool calls.
To our knowledge, prior work does not evaluate MCP-style agents for hardware design workflows involving persistent cross-task state, invalid requests, or multi-server tool contexts.

\noindent\textbf{Agent configuration.}
Tool-calling performance depends not only on the model but also on the surrounding system design.
Orchestration strategy matters: AgentArch~\cite{bogavelli2025agentarch} shows that architectural preferences are model-dependent, with different designs benefiting different model scales.
Prompt structure significantly affects tool-use behaviour, with structured and role-based prompts improving compliance~\cite{he2024does,zhang2024sprig}.
Tool-description quality shapes tool selection and argument accuracy~\cite{naturaltools}, while long contexts can degrade performance over extended sessions~\cite{liu2024lost,levy2024same}.
Although these factors have been studied independently, their combined effect in stateful, dependency-ordered environments remains underexplored.

\section{Experimental Setup}
\label{sec:setup}

\subsection{MCP Server}

We implement a Model Context Protocol (MCP) server that reproduces the state, data model, and dependency constraints of a proprietary hardware design tool through 14 callable tools (Table~\ref{tab:tools}), where later calls may depend on earlier ones.

\begin{table}[!t]
\centering
\small
\setlength{\tabcolsep}{4pt}
\begin{tabular}{lp{5.2cm}}
\textbf{Group} & \textbf{Tools} \\
\midrule
Adder    & \texttt{create\_component}, \texttt{add\_signal\_port},
           \texttt{add\_transactional\_port},
           \texttt{add\_user\_parameter},
           \texttt{add\_generator\_parameter},
           \texttt{add\_subcomponent},
           \texttt{add\_subcomponent\_connection} \\[3pt]
Getter   & \texttt{get\_component\_details},
           \texttt{get\_element},
           \texttt{list\_components} \\[3pt]
Other    & \texttt{delete\_element}, \texttt{delete\_component},
           \texttt{save\_component}, \texttt{load\_component} \\
\end{tabular}
\caption{MCP server tool set.}
\label{tab:tools}
\end{table}

\subsection{Models}

We evaluate seven open-source models, all running locally through Ollama~\cite{ollama} as 4-bit quantised variants: Llama~3.1~8B~\cite{grattafiori2024llama}, Gemma~4~E4B, Gemma~4~26B, Gemma~4~31B~\cite{gemma4}, Qwen~3.5~27B~\cite{qwen35}, Qwen~3.6~27B~\cite{qwen36}, and GPT-OSS~20B~\cite{agarwal2025gpt}.
All runs use each model's default Ollama sampling temperature reflecting an out-of-the-box local deployment setting;
Section~\ref{sec:stability} reports sensitivity checks at lower temperatures.

\subsection{Agent Design}
\label{sec:config}

\noindent\textbf{Agent architecture.}
We compare two architectures: \textbf{ReAct}~\cite{yao2022react}, a single-agent loop issuing one tool call per turn, and \textbf{Plan-and-Act}~\cite{erdogan2025plan}, a multi-agent design in which a planner decomposes the request, a validator checks the plan, and independent ReAct workers execute the steps (sequence diagrams in Appendix~\ref{app:arch}).

\noindent\textbf{System prompt.}
Four ReAct variants are evaluated: \textbf{none} (tool schemas only), \textbf{basic} (role and behavioural constraints), \textbf{MD} (the same content in Markdown with structured headers), and \textbf{fewshot} (\textbf{basic} extended with worked task--call examples).
For Plan-and-Act, planner and validator each have \textbf{basic} and \textbf{structured} variants; the worker uses a fixed prompt.
Full prompt texts are provided in Appendix~\ref{app:prompts}.

\noindent\textbf{Tool description format.}
\textbf{Comprehensive} descriptions specify each tool's purpose, parameter semantics, constraints, and failure conditions.
\textbf{Minimal} descriptions reduce each tool to a single sentence, saving approximately 2{,}000 tokens across the 14 tools.

\noindent\textbf{History scope.}
Under \textbf{run} scope, the full interaction history accumulates across all tasks in a session.
Under \textbf{task} scope, each task starts with a fresh context.

\begin{figure*}[t]
  \centering
  \includegraphics[width=\textwidth]{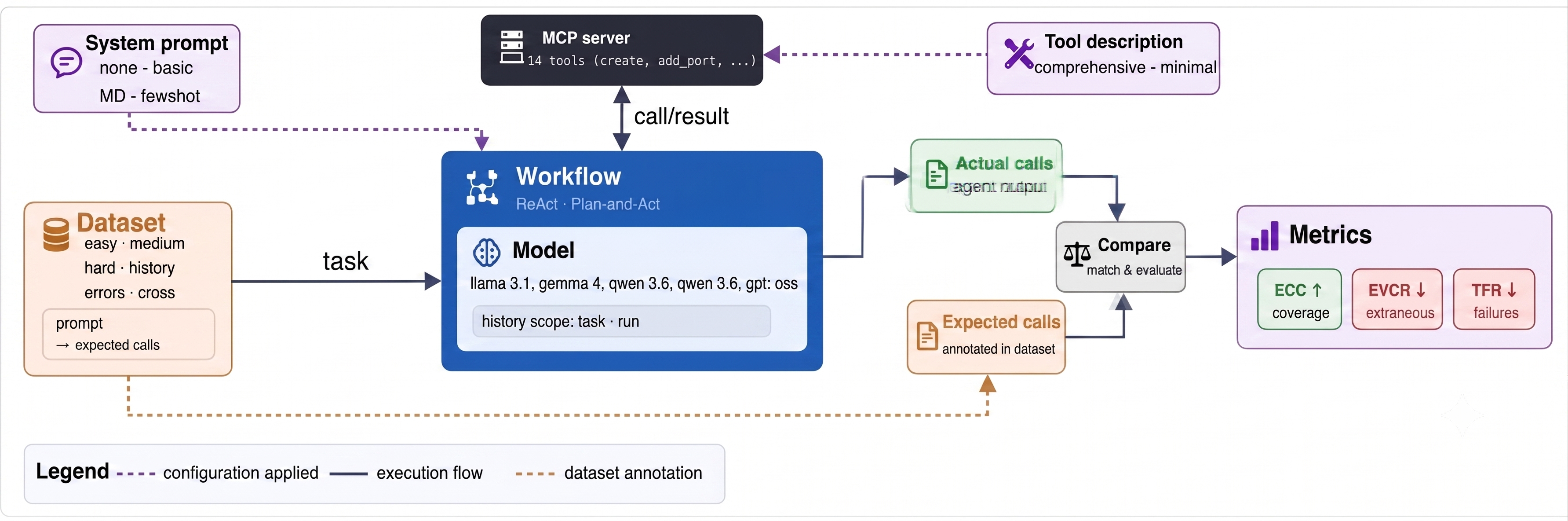}
  \caption{Evaluation pipeline.
Each task prompt is submitted to the agent, which issues tool calls to the MCP server and receives results.
The produced call sequence is compared against annotated expected calls to compute the metrics of Section~\ref{sec:metrics}.}
  \label{fig:pipeline}
\end{figure*}

\subsection{Benchmark}
\label{sec:datasets}

We construct an expert-informed benchmark comprising six core task suites---Easy, Medium, Hard, History, Errors, and Cross---and two multi-server suites (Easy-Noise, Hard-Noise).
Professional users of the design tool identified recurring hardware design operations, reviewed the task categories, and validated expected call sequences by executing them against the server.
Task prompts reflect routine component-development operations.
The full benchmark cannot be released because it encodes proprietary workflow details;
representative examples are provided in Appendix~\ref{app:examples}.

\noindent\textbf{Independent tasks.}
(Appendix~\ref{app:ds_easy}--\ref{app:ds_hard}) Easy, Medium, and Hard contain 40 self-contained tasks each and isolate the effect of task length: Easy requires 1 expected call, Medium requires 2, and Hard requires 3--5 calls with dependencies.
Each prompt provides all required identifiers, such as the full VLNV of a component or the names of the ports being connected, so no information from prior turns is needed.

\noindent\textbf{History.}
(Appendix~\ref{app:ds_history}) History contains 40 tasks whose prompts omit entity identifiers and instead refer to prior tasks, for example, ``Add the same port to the component created in the previous task.''
The agent must recover the relevant component and port information from earlier turns before acting.

\noindent\textbf{Errors.}
(Appendix~\ref{app:ds_errors_misspelled}) Errors contains 40 tasks.
Twenty contain heavily misspelled prompts with the same intended content as Hard tasks, enabling a paired comparison of prompt noise.
The remaining 20 describe requests for which the correct response is to make no tool call.

\noindent\textbf{Cross.}
(Appendix~\ref{app:ds_cross}) Cross contains 40 tasks that combine dependency chains, cross-task references, pattern expansion, and error detection within a single multi-task session.
For example, early tasks create components \texttt{sub1} and \texttt{sub2} with identical port sets;
a later task asks to ``create \texttt{sub3} like the others and wire it the same way'';
and a final task requests an operation on a component that was never created, which the agent must reject rather than execute.

\noindent\textbf{Noisy context.}
(Appendix~\ref{app:ds_noise}) Noisy context contains two 60-task suites evaluated on three models.
Each suite interleaves one external-server task after every two system design tasks while exposing the design server alongside GDB debugging\footnote{\nolinkurl{https://github.com/signal-slot/mcp-gdb}} and Git version-control\footnote{\nolinkurl{https://github.com/modelcontextprotocol/servers/tree/main/src/git}} MCP servers.
\textbf{Easy-Noise} uses the 40 Easy tasks; \textbf{Hard-Noise} uses the 40 Hard tasks.
Because the system design tasks are unchanged, the clean suites provide a paired baseline for measuring routing errors and long-session effects.

\subsection{Evaluation Metrics}
\label{sec:metrics}

Evaluating stateful agents requires call-level metrics rather than textual output assessment~\cite{yao2024tau,gao2025mcp}.
We define four complementary metrics, interpreted jointly.

\noindent\textbf{Expected Call Coverage (ECC, $\uparrow$)} measures how much of the annotated work was completed:
\begin{equation}
  \mathrm{ECC} =
  \frac{|\text{correctly executed expected calls}|}{|\text{expected calls}|}.
\end{equation}
A call matches an expected call if it succeeds with the expected tool name and all expected argument values;
each expected call is consumed at most once.
Matching is order-insensitive because dependency-order violations are enforced by the server, which rejects invalid calls; such failures are accounted for in TFR.
Tasks with an empty expected call set are excluded from ECC aggregation.

\noindent\textbf{Extraneous Valid Call Ratio (EVCR, $\downarrow$)} measures over-generation, i.e., successful calls that were not requested:
\begin{equation}
  \mathrm{EVCR} =
  \frac{|\text{successful non-expected calls}|}
       {|\text{successful calls}|}
\end{equation}
If an agent issues no successful calls, EVCR is defined as $0$ by convention.

\noindent\textbf{Tool Failure Rate (TFR, $\downarrow$)} measures the fraction of calls rejected by the server:
\begin{equation}
  \mathrm{TFR} =
  \frac{|\text{failed calls}|}{|\text{total calls}|}
\end{equation}
If an agent issues no calls, TFR is defined as $0$.

\noindent\textbf{No-Call Accuracy (NCA, $\uparrow$)} measures correct abstention on tasks with an empty expected call set:
\begin{equation}
  \mathrm{NCA} =
  \frac{|\text{empty-expected tasks with no tool call}|}
       {|\text{empty-expected tasks}|}
\end{equation}

\noindent\textbf{Evaluation protocol.}
Tasks are presented sequentially.
After each task, we remove all mutations produced by the agent and restore the server to the canonical state for the next task by replaying the expected sequence of calls up to that task.
This prevents one agent error from making later tasks impossible while preserving the intended state dependencies in History and Cross.
Figure~\ref{fig:pipeline} illustrates the pipeline.
For ReAct, all configuration combinations are evaluated on the six core suites (Easy through Cross); the noisy-context suites are evaluated on three representative models.
\begin{figure}[h!]
  \centering
  \includegraphics[width=\columnwidth]{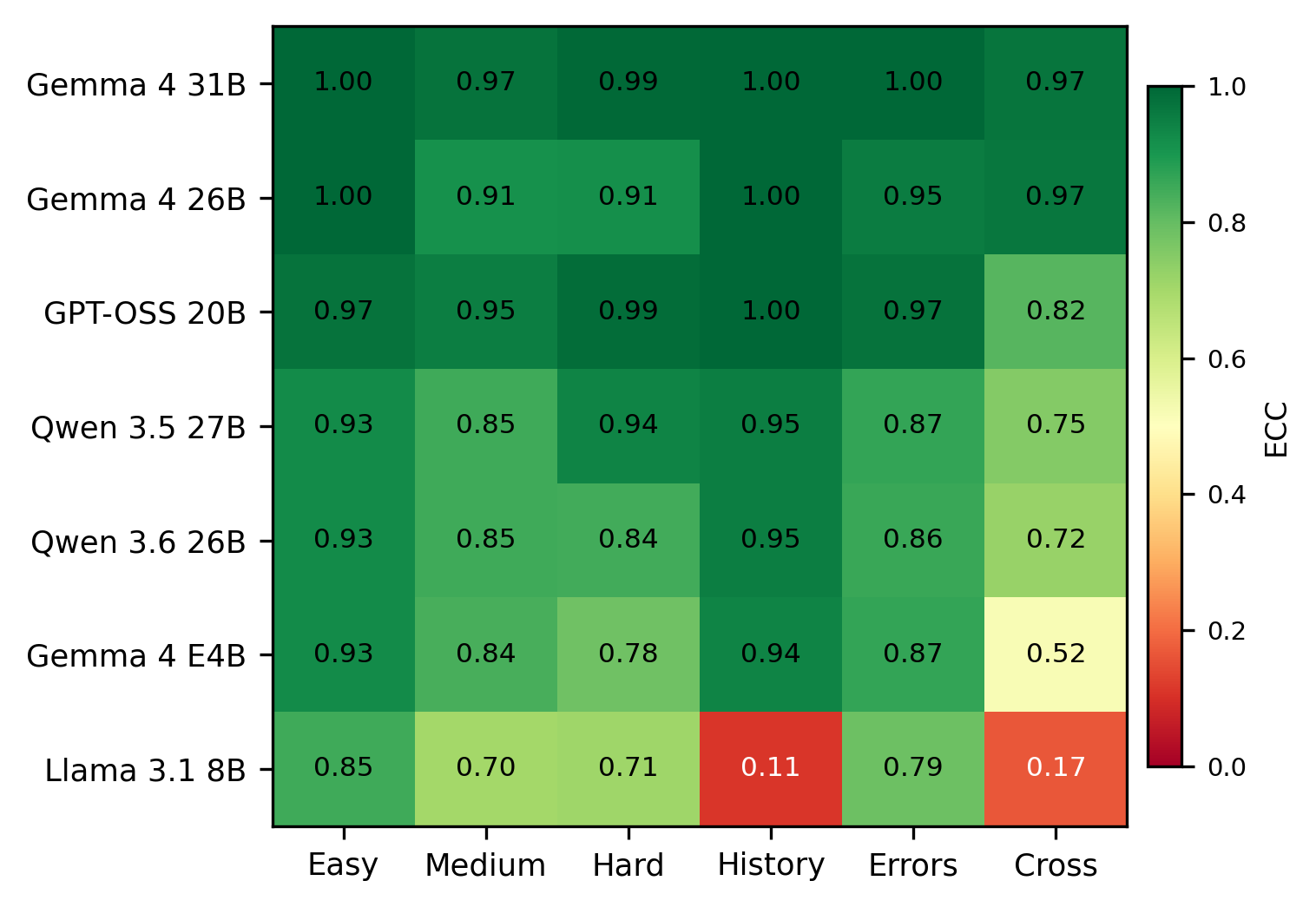}
  \caption{Best-configuration ECC per model and task suite under ReAct.}
  \label{fig:heatmap}
\end{figure}

\section{Results}
\label{sec:results}

\subsection{Can Hardware Design Be Automated?}
\label{sec:overview}

Table~\ref{tab:overview} reports, for each model, the ReAct configuration that maximises average ECC across the six core task suites, together with the resulting EVCR and TFR.
These numbers should be interpreted as best-observed performance under configuration search, not as out-of-the-box model performance.
Plan-and-Act is considered separately below.

\begin{table}[h!]
\centering
\small
\setlength{\tabcolsep}{3pt}
\begin{tabular}{lcccc}
& \textbf{Config} & \textbf{ECC}\,$\uparrow$ & \textbf{EVCR}\,$\downarrow$ & \textbf{TFR}\,$\downarrow$ \\
\midrule
Gemma~4~31B  & R/MD/C   & 0.990 & 0.015 & 0.011 \\
Gemma~4~26B  & R/MD/C   & 0.958 & 0.021 & 0.061 \\
GPT-OSS~20B  & R/none/C & 0.951 & 0.073 & 0.049 \\
Qwen~3.5~27B & R/MD/M   & 0.880 & 0.036 & 0.048 \\
Qwen~3.6~27B & R/MD/M   & 0.858 & 0.086 & 0.065 \\
Gemma~4~E4B  & R/MD/C   & 0.811 & 0.023 & 0.058 \\
Llama~3.1~8B & T/fs/C   & 0.554 & 0.064 & 0.353 \\
\end{tabular}
\caption{Best single ReAct configuration per model, averaged over 
the six core task suites and ordered by ECC.
Config notation: history / prompt / tools, where R\,=\,run, T\,=\,task, C\,=\,comprehensive, M\,=\,minimal, fs\,=\,few-shot.}
\label{tab:overview}
\end{table}

The aggregate hides a strong interaction between model and task structure.
Figure~\ref{fig:heatmap} breaks down ECC by task suite: the model gap is small on simple tasks and widens as tasks require more state recovery and dependency management.
Gemma~4~E4B stays within $6$--$14\%$ of Gemma~4~31B on five of six task suites, but on Cross the gap widens to $47\%$ ($0.517$ vs.\ $0.972$).
The same pattern appears within a single model: Llama~3.1~8B reaches $0.753$ average ECC on the independent task suites (Easy, Medium, Hard), but only $0.139$ on the two suites that require carrying state across tasks (History, Cross), an $82\%$ drop.

Best configuration is itself workload-dependent.
Table~\ref{tab:overview} reports the configuration that maximises average ECC across all six core suites, but for Llama~3.1~8B this global optimum gives only $0.113$ ECC on History, whereas the History-specific optimum reaches $0.538$.
Even the globally best configuration is not reached automatically: under a poor configuration, Gemma~4~E4B drops from $0.811$ to $0.168$ ECC on the same tasks.
Thus, model choice sets the broad performance range, but configuration determines whether a model reaches that range in practice.

Figure~\ref{fig:heatmap} omits two important failure modes.
Table~\ref{tab:nocall} reports both.
First, the Errors suite includes 20 tasks with no correct tool call;
for these tasks ECC is undefined, so we report NCA, EVCR, and TFR.
Gemma~4~31B is nearly perfect, but other models either call unnecessary valid tools, issue failing calls, or both.
Second, co-locating the design server with GDB and Git servers leaves ECC largely unchanged at matched configuration for the three models tested, but raises EVCR for all three, suggesting that a larger tool context mainly increases extraneous routing errors rather than reducing coverage.

\begin{table}[h!]
\centering
\small
\setlength{\tabcolsep}{3.5pt}
\begin{tabular}{lccccc}
& \multicolumn{3}{c}{\textbf{No-call (Errors)}} & \multicolumn{2}{c}{\textbf{Noise (Hard)}} \\
& NCA & EVCR & TFR & clean & noisy \\
\midrule
Gemma~4~31B  & {0.80} & 0.00 & 0.05 & ---  & ---  \\
Gemma~4~26B  & {0.40} & 0.05 & 0.55 & 0.00 & 0.16 \\
GPT-OSS~20B  & {0.25} & 0.58 & 0.13 & ---  & ---  \\
Qwen~3.5~27B & {0.70} & 0.15 & 0.17 & ---  & ---  \\
Qwen~3.6~27B & {0.45} & 0.20 & 0.37 & ---  & ---  \\
Gemma~4~E4B  & {0.65} & 0.10 & 0.25 & 0.00 & 0.05 \\
Llama~3.1~8B & {0.00} & 0.25 & 
0.82 & 0.12 & 0.22 \\
\end{tabular}
\caption{NCA, EVCR, and TFR on the 20 no-call tasks in Errors, at each model's best configuration.
Right: EVCR on Hard vs.\ Hard-Noise under task scope and matched configuration for the three models tested.
ECC changes are small and omitted for space.}
\label{tab:nocall}
\end{table}

Model and task structure are the two primary factors.
The remaining results ask which agent-configuration choices still matter once those two are fixed.

\subsection{How Much Does Agent Configuration Matter?}
\label{sec:config_effects}

\noindent\textbf{System prompt.}
Markdown-formatted instructions (MD) are the best prompt for five of the seven models, although the margin over none/basic is generally small (under $0.04$ ECC).
The main exception is few-shot prompting.
For five models, few-shot changes ECC by at most $0.04$ on average and often reduces EVCR and TFR;
for Gemma~4~31B and Gemma~4~E4B, however, it causes severe inaction.
Gemma~4~31B drops from $0.956$ ECC with the best non-few-shot prompt to $0.571$ with few-shot, and Gemma~4~E4B drops from $0.731$ to $0.179$.
In both cases EVCR and TFR fall together with ECC, indicating that the models stop acting rather than acting incorrectly.
In this benchmark, prompt engineering has an asymmetric risk profile: gains over simpler prompts are modest for most models, while a poorly matched few-shot prompt can be catastrophic.
Detailed prompt ablations are reported in Appendix~\ref{app:prompt_results}.

\noindent\textbf{Tool descriptions.}
Comprehensive tool descriptions provide the most consistent configuration benefit.
Holding all other choices fixed, switching to minimal descriptions raises TFR for every model, roughly doubling it for most.
The effect on ECC is less uniform, but the reliability gain suggests that parameter semantics, constraints, and failure conditions help models construct valid calls.
Across the $42$ model--task-suite combinations in Figure~\ref{fig:heatmap}, the configuration that maximises ECC on that suite uses comprehensive descriptions in $35$ cases ($83\%$).
Each model's globally best configuration also uses comprehensive descriptions in five of seven cases.
Full tool-description ablations are reported in Appendix~\ref{app:tool_description_results}.

\noindent\textbf{Context and history scope.}
On independent tasks, history scope has little effect for six of the seven models: ECC shifts by at most $0.031$.
Llama~3.1~8B is the exception.
On Easy, Medium, and Hard, cumulative history cuts its ECC from $0.667$ to $0.192$, a $71\%$ drop, while TFR also falls from $0.194$ to $0.070$.
This pattern indicates silence rather than more frequent invalid calls.
For constrained models, retaining irrelevant history can therefore be actively harmful.
Full history-scope results are reported in Appendix~\ref{app:history_results}.

\noindent\textbf{Architecture.}
Table~\ref{tab:arch} compares ReAct with Plan-and-Act using Gemma~4~26B as planner.
This is a pipeline-level comparison: Plan-and-Act can improve weak workers partly by delegating decomposition to a stronger model.
With Llama~3.1~8B as worker, decomposition raises average ECC from $0.554$ to $0.718$, with the largest gains on History and Cross.
With Gemma~4~26B as worker on the same suites, decomposition does not help and reduces coverage, suggesting that strong single-agent workers benefit from retaining full session context.
On longer Hard-Noise sessions under cumulative history, however, Plan-and-Act recovers coverage from $0.858$ to $0.950$ at the cost of higher EVCR.
Thus, multi-agent decomposition is most useful when the worker is weak or the session is long enough for single-agent context accumulation to become a bottleneck.

\begin{table}[h!]
\centering
\small
\setlength{\tabcolsep}{4pt}
\begin{tabular}{llcc}
\textbf{Worker} & \textbf{Suite} & \textbf{ReAct} & \textbf{Plan-and-Act} \\
\midrule
Llama~3.1~8B & Avg.\ (6 core) & 0.554 & \textbf{0.718} \\
Llama~3.1~8B & History        & 0.113 & \textbf{0.650} \\
Llama~3.1~8B & Cross          & 0.166 & \textbf{0.450} \\
\midrule
Gemma~4~26B  & Hard-Noise (60) & 0.858 & \textbf{0.950} \\
\end{tabular}
\caption{ECC under ReAct vs.\ Plan-and-Act, at each worker's best
configuration (averaged over the six core suites).
The
Hard-Noise row uses a fixed run-scope configuration, isolating the
effect of session length.
}
\label{tab:arch}
\end{table}

\subsection{Stability}
\label{sec:stability}
All main experiments use each model's default temperature (1.0, except
0.8 for Llama 3.1 8B). Re-running all configurations on three
representative datasets across temperatures and repeating a subset of
configurations under identical conditions confirms that the main
effects are not artefacts of sampling. Across temperatures 0, 0.5, and
1.0, model rankings are preserved and ECC shifts remain small; for
Gemma 4 26B, ECC ranges from 0.881 to 0.896. Run-to-run
variance is low for Gemma 4 26B and moderate for weaker models; the
largest outlier corresponds to the few-shot collapse described above.
Full stability results are reported in Appendix~\ref{app:stability}.

\section{Conclusion}

We evaluated whether LLM agents can automate expert-defined hardware design tasks through MCP tool calling. On our benchmark, the strongest open-source models achieve near-perfect expected-call coverage under their best configuration, including on multi-step sessions with implicit state and co-located GDB and Git servers.

Model choice primarily determines the achievable performance range, while system configuration determines whether that performance is actually reached in practice.

Our results suggest five deployment practices: test prompts on the target model; use comprehensive tool descriptions; benchmark models on workload-representative tasks; manage context for constrained models; and reserve multi-agent decomposition for weak workers or long sessions.

These findings indicate that local LLM agents are practical for structured, stateful hardware component-editing workflows when the tool-calling pipeline is appropriately configured.

\newpage
\section{Limitations}
\label{sec:limitations}

The main limitation is the scope of our context-management exploration. We tested only the binary choice between per-task and cumulative history; mechanisms such as a variable memory window—summarising or discarding older turns rather than retaining or dropping all of them—could potentially improve weaker models on session-dependent tasks, and our results hint at this but do not test it. Relatedly, our session-dependent datasets are limited to tens of tasks; longer sessions might reveal additional context-management effects for models beyond Llama 3.1 8B, which was the only model showing sensitivity to history scope at this session length.

Second, EVCR counts extraneous successful calls but does not weight their severity: an unnecessary user parameter and an incorrect subcomponent connection contribute equally to the ratio, though the latter has far greater impact on design correctness. A severity-weighted variant is left for future work.

Third, we do not systematically measure response latency. Local inference time scales with model size, and a latency-sensitive deployment might reasonably prefer a smaller model at some cost in coverage—a trade-off our metrics do not capture.

Finally, the MCP server mirrors the data model and dependency rules of the proprietary design tool but does not call its production APIs; the results therefore characterise agent behaviour against a faithful replica, and some gap with production behaviour should be expected. Neither the server implementation nor the benchmark tasks can be released, as both encode proprietary component specifications and workflow details; Appendix~\ref{app:examples} provides representative task examples to support methodological reproduction.

\section*{Acknowledgments}

We thank Huawei for the opportunity to carry out this work during Leonardo’s internship. We are especially grateful to Rahul Setia, Leonardo’s supervisor at Huawei, for his guidance throughout the work, and to Johan Hokfelt for helping make this submission possible.

\bibliography{references}

\clearpage
\appendix
\raggedbottom
\onecolumn

\section{Agent Architecture Diagrams}
\label{app:arch}

\subsection{ReAct}

\begin{figure}[H]
  \centering
  \includegraphics[width=\textwidth]{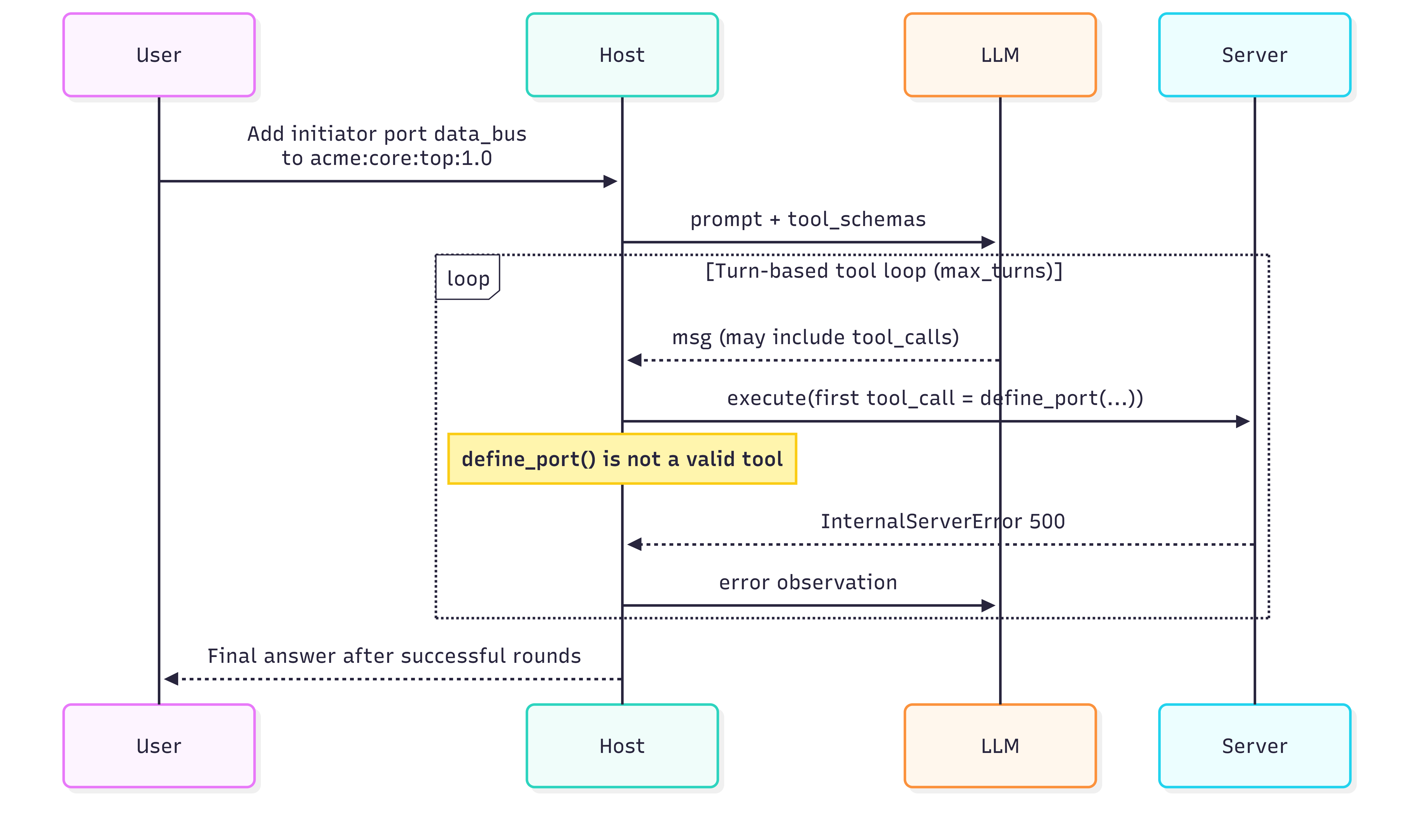}
  \caption{ReAct sequence diagram. The host submits the task prompt and
  tool schemas to the LLM.
  Within a turn-bounded loop, the LLM returns a
  response that may include a tool call;
  the host executes the first call,
  appends the result, and repeats.
  The loop terminates when no further tool
  call is produced or the turn limit is reached.}
  \label{fig:react_diag}
\end{figure}

\subsection{Plan-and-Act}

\begin{figure}[H]
  \centering
  \includegraphics[width=\textwidth]{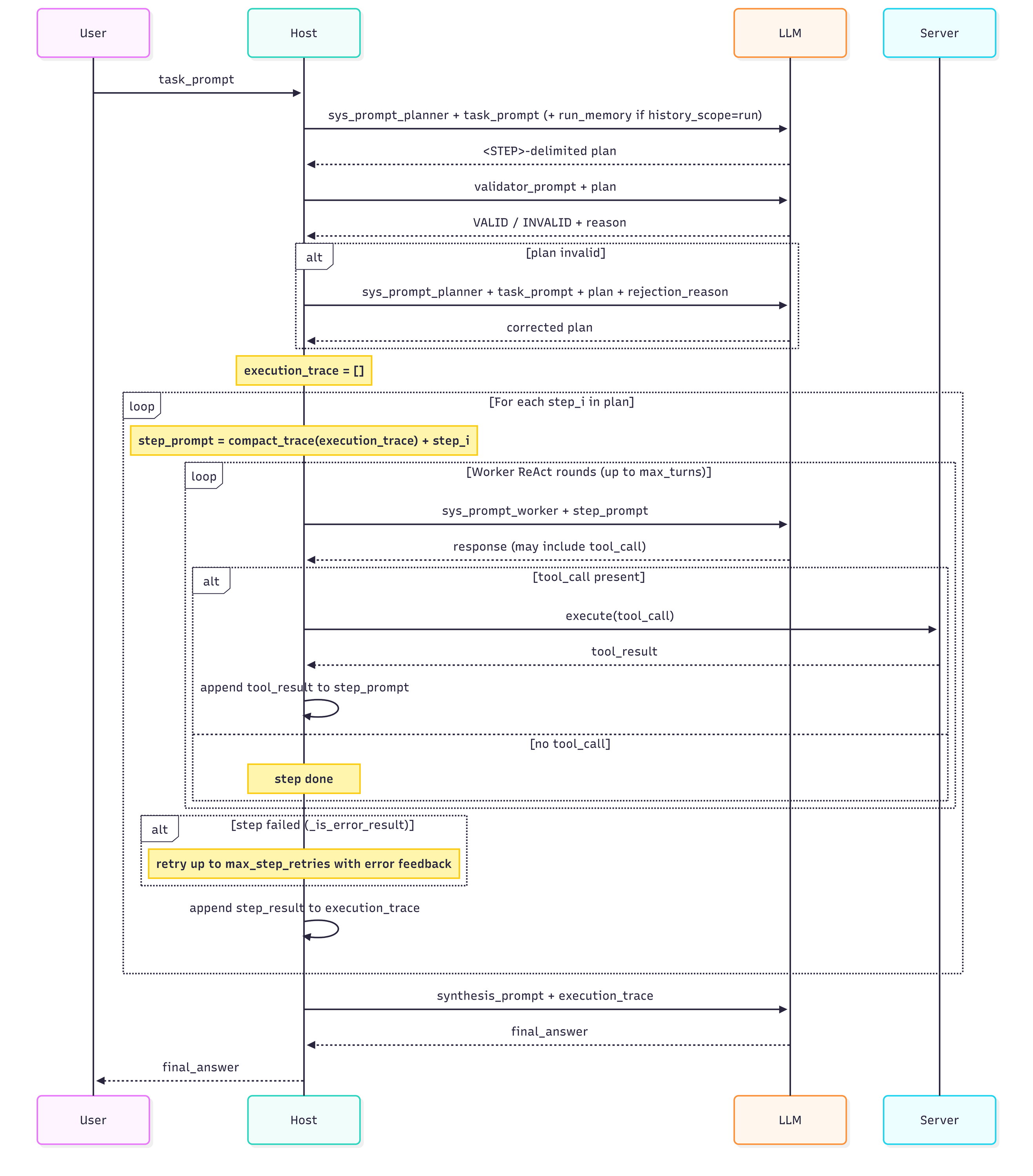}
  \caption{Plan-and-Act sequence diagram. The planner produces a
  \texttt{<STEP>}-delimited plan;
  the validator checks it with one replan
  permitted. Each step is executed by an independent ReAct worker with a
  compact trace of prior results.
  A synthesis call assembles the final
  answer.}
  \label{fig:planact_diag}
\end{figure}

\twocolumn
\section{Task Examples}
\label{app:examples}

\subsection{Easy}\label{app:ds_easy}
\begin{taskcard}{\textbf{EASY\_001}}
\textit{``Create a component with vendor acme, library core, name top, and version 1.0.''}
\tcblower
create\_component(acme, core, top, 1.0)
\end{taskcard}
 
\begin{taskcard}{\textbf{EASY\_002}}
\textit{``Add an input signal port named clk of width 1 to component acme:core:top:1.0.''}
\tcblower
\textnormal{[on acme:core:top:1.0]} \\ add\_signal\_port(clk, in, w=1)
\end{taskcard}
 
\begin{taskcard}{\textbf{EASY\_003}}
\textit{``Add an initiator transactional port named tx to component acme:core:top:1.0.''}
\tcblower
\textnormal{[on acme:core:top:1.0]} \\ add\_transactional\_port(tx, initiator)
\end{taskcard}
 
\begin{taskcard}{\textbf{EASY\_004}}
\textit{``Add an output signal port named data\_out of width 8 to component acme:core:top:1.0.''}
\tcblower
\textnormal{[on acme:core:top:1.0]} \\ add\_signal\_port(data\_out, out, w=8)
\end{taskcard}
 
\begin{taskcard}{\textbf{EASY\_005}}
\textit{``Add a user parameter named DATA\_WIDTH of type integer with default value 32 to component acme:core:top:1.0.''}
\tcblower
\textnormal{[on acme:core:top:1.0]} \\ add\_user\_parameter(DATA\_WIDTH, integer, =32)
\end{taskcard}
 
\subsection{Medium}\label{app:ds_medium}
\begin{taskcard}{\textbf{MED\_001}}
\textit{``Create a component riscv\_core (vendor openhw, lib cpu, v 1.0) and add a transactional initiator port named data\_bus.''}
\tcblower
create\_component(openhw, cpu, riscv\_core, 1.0) \\ add\_transactional\_port(data\_bus, initiator)
\end{taskcard}
 
\begin{taskcard}{\textbf{MED\_002}}
\textit{``The riscv\_core also needs a clock input.
Add a 1-bit input signal port named clk, then save.''}
\tcblower
\textnormal{[on openhw:cpu:riscv\_core:1.0]} \\ add\_signal\_port(clk, in, w=1) \\ save\_component
\end{taskcard}
 
\begin{taskcard}{\textbf{MED\_003}}
\textit{``Create a component sram (vendor openhw, lib mem, v 1.0) and add a transactional target port slave\_port with a memory implementation called mem of 1024 bytes.''}
\tcblower
create\_component(openhw, mem, sram, 1.0) \\ add\_transactional\_port(slave\_port, target, impl=memory, mem=mem, bytes=1024)
\end{taskcard}
 
\begin{taskcard}{\textbf{MED\_004}}
\textit{``Complete the sram component by adding a 1-bit input signal named clk, then save it.''}
\tcblower
\textnormal{[on openhw:mem:sram:1.0]} \\ add\_signal\_port(clk, in, w=1) \\ save\_component
\end{taskcard}
 
\subsection{Hard}\label{app:ds_hard}
\begin{taskcard}{\textbf{HARD\_001}}
\textit{``Let's start our SoC design by creating the main CPU.
Please define a component named riscv\_tile from vendor acme, library ip\_cores, version 2.1.0.
Add an initiator transactional port called m\_axi for memory access and a signal input port for the clk with a width of 1 and multiplicity of 1. Finally, save the component to disk.''}
\tcblower
create\_component(acme, ip\_cores, riscv\_tile, 2.1.0) \\ add\_transactional\_port(m\_axi, initiator) \\ add\_signal\_port(clk, in, w=1) \\ save\_component
\end{taskcard}
 
\begin{taskcard}{\textbf{HARD\_002}}
\textit{``We need a memory controller now.
Create sram\_ctrl (vendor: acme, lib: mem, v: 1.0.0). It needs a target transactional port named s\_axi using a memory implementation called mem of 30 bytes.
Also, add a 1-bit input signal port for rst\_n and a 1-bit input clk.
Save it when done.''}
\tcblower
create\_component(acme, mem, sram\_ctrl, 1.0.0) \\ add\_transactional\_port(s\_axi, target, impl=memory, mem=mem, bytes=30) \\ add\_signal\_port(rst\_n, in, w=1) \\ add\_signal\_port(clk, in, w=1) \\ save\_component
\end{taskcard}
 
\begin{taskcard}{\textbf{HARD\_003}}
\textit{``Let's define a system interconnect component named axi\_interconnect (vendor: acme, lib: bus, v: 1.1).
Add one initiator transactional port m0\_port and two target transactional ports s0\_port and s1\_port with user\_defined implementation.
Then save the component.''}
\tcblower
create\_component(acme, bus, axi\_interconnect, 1.1) \\ add\_transactional\_port(m0\_port, initiator) \\ add\_transactional\_port(s0\_port, target, impl=user\_defined) \\ add\_transactional\_port(s1\_port, target, impl=user\_defined) \\ save\_component
\end{taskcard}
 
\subsection{History}\label{app:ds_history}
Tasks run in session order;
later tasks omit identifiers and must recover them from earlier turns.\smallskip
 
\begin{taskcard}{\textbf{HIST\_001}}
\textit{``Create a RISC-V core openhw:cpu:riscv\_v5:1.0 and add a transactional initiator port data\_master.''}
\tcblower
create\_component(openhw, cpu, riscv\_v5, 1.0) \\ add\_transactional\_port(data\_master, initiator)
\end{taskcard}
 
\begin{taskcard}{\textbf{HIST\_002}}
\textit{``Add another transactional initiator port called instr\_master.''}
\\[1pt]\textcolor{black!60}{\small\textit{Note:} no id given;
recover openhw:cpu:riscv\_v5:1.0 from HIST\_001.}
\tcblower
\textnormal{[on openhw:cpu:riscv\_v5:1.0]} \\ add\_transactional\_port(instr\_master, initiator)
\end{taskcard}
 
\begin{taskcard}{\textbf{HIST\_003}}
\textit{``Create system\_sram (generic, mem, 2.1) with a target port mem\_port, memory mem, 65536 bytes.''}
\tcblower
create\_component(generic, mem, system\_sram, 2.1) \\ add\_transactional\_port(mem\_port, target, impl=memory, mem=mem, bytes=65536)
\end{taskcard}
 
\begin{taskcard}{\textbf{HIST\_004}}
\textit{``Create axi\_interconnect (generic, bus, 1.0) and add a user\_defined target port slave\_0.''}
\tcblower
create\_component(generic, bus, axi\_interconnect, 1.0) \\ add\_transactional\_port(slave\_0, target, impl=user\_defined)
\end{taskcard}
 
\begin{taskcard}{\textbf{HIST\_005}}
\textit{``Go back to the RISC-V core.
Add a 1-bit input clock and a 1-bit input reset\_n.''}
\\[1pt]\textcolor{black!60}{\small\textit{Note:} "the RISC-V core" = openhw:cpu:riscv\_v5:1.0 from HIST\_001.}
\tcblower
\textnormal{[on openhw:cpu:riscv\_v5:1.0]} \\ add\_signal\_port(clock, in, w=1) \\ add\_signal\_port(reset\_n, in, w=1)
\end{taskcard}
 
\subsection{Errors --- misspelled vs.\ clean}\label{app:ds_errors_misspelled}
Each misspelled Errors prompt mirrors a Hard prompt with heavy orthographic noise but identical intent.
The pairs below share the same expected calls, isolating the effect of prompt noise.\smallskip
 
\begin{tcolorbox}[taskcardbase, breakable, width=\columnwidth, title={HARD\_001  vs  ERR\_001}]
\textnormal{\itshape Clean (Hard):}\\[2pt]
``\textit{Let's start our SoC design by creating the main CPU.
Please define a component named riscv\_tile from vendor acme, library ip\_cores, version 2.1.0.
Add an initiator transactional port called m\_axi and a signal input port for clk with width 1 and multiplicity 1. Finally, save the component to disk.}''

\smallskip
\textnormal{\itshape Misspelled (Errors):}\\[2pt]
``\textit{Let's start our SoC design by cretaing the main CPU.
Please defnie a compnnt named riscv\_tile from vendor acme, library ip\_cores, version 2.1.0.
Add an initator transactonal prt called m\_axi and a siganl inpt prt for clk with width 1 and multiplicty 1. Finally, saev the compnnt to disk.}''

\tcblower
\textnormal{\footnotesize Identical expected calls:}\\[1pt]
create\_component(acme, ip\_cores, riscv\_tile, 2.1.0) \\ add\_transactional\_port(m\_axi, initiator) \\ add\_signal\_port(clk, in, w=1) \\ save\_component
\end{tcolorbox}
 
\begin{tcolorbox}[taskcardbase, breakable, width=\columnwidth, title={HARD\_002  vs  ERR\_002}]
\textnormal{\itshape Clean (Hard):}\\[2pt]
``\textit{We need a memory controller now.
Create sram\_ctrl (acme, mem, 1.0.0). It needs a target transactional port named s\_axi using a memory implementation called mem of 30 bytes.
Also add a 1-bit input signal port for rst\_n and a 1-bit input clk.
Save it when done.}''

\smallskip
\textnormal{\itshape Misspelled (Errors):}\\[2pt]
``\textit{We need a memroy controller now. Cretae sram\_ctrl (acme, mem, 1.0.0).
It needs a trgt transactinal prt named s\_axi usng a memroy implmentation called mem of 30 bytes.
Also add a 1-bit input siganl prt for rst\_n and a 1-bit input clk.
Saev it when done.}''

\tcblower
\textnormal{\footnotesize Identical expected calls:}\\[1pt]
create\_component(acme, mem, sram\_ctrl, 1.0.0) \\ add\_transactional\_port(s\_axi, target, impl=memory, mem=mem, bytes=30) \\ add\_signal\_port(rst\_n, in, w=1) \\ add\_signal\_port(clk, in, w=1) \\ save\_component
\end{tcolorbox}
 
\begin{tcolorbox}[taskcardbase, breakable, width=\columnwidth, title={HARD\_003  vs  ERR\_003}]
\textnormal{\itshape Clean (Hard):}\\[2pt]
``\textit{Let's define a system interconnect component named axi\_interconnect (acme, bus, 1.1).
Add one initiator transactional port m0\_port and two target transactional ports s0\_port and s1\_port with user\_defined implementation.
Then save the component.}''

\smallskip
\textnormal{\itshape Misspelled (Errors):}\\[2pt]
``\textit{Let's defnie a system interconect compnent axi\_interconnect (acme, bus, 1.1).
Add one initator transactonal prt m0\_port and two trgt transactonal prts s0\_port and s1\_port with user\_defned implmentation.
Then saev the compnent.}''

\tcblower
\textnormal{\footnotesize Identical expected calls:}\\[1pt]
create\_component(acme, bus, axi\_interconnect, 1.1) \\ add\_transactional\_port(m0\_port, initiator) \\ add\_transactional\_port(s0\_port, target, impl=user\_defined) \\ add\_transactional\_port(s1\_port, target, impl=user\_defined) \\ save\_component
\end{tcolorbox}

\FloatBarrier

\subsection{Errors --- no-call}\label{app:ds_errors_nocall}
The correct behaviour is to issue no call and explain why.\smallskip
 
\begin{taskcardnc}{\textbf{ERR\_004}}
\textit{``Create a new component for our DMA block with vendor acme, library interconnect, name dma.''}
\tcblower
\textnormal{\itshape no call --- }version is required but missing;
cannot be inferred.
\end{taskcardnc}
 
\begin{taskcardnc}{\textbf{ERR\_005}}
\textit{``Add a signal port called irq to component acme:cpu:core:1.0.''}
\tcblower
\textnormal{\itshape no call --- }direction and width missing;
component also absent from state.
\end{taskcardnc}
 
\begin{taskcardnc}{\textbf{ERR\_006}}
\textit{``On component acme:mem:sram\_ctrl:2.1, add a transactional port named cfg as a target backed by memory.''}
\tcblower
\textnormal{\itshape no call --- }memory implementation needs a memory name and size;
neither given.
\end{taskcardnc}
 
\begin{taskcardnc}{\textbf{ERR\_007}}
\textit{``Save the component.''}
\tcblower
\textnormal{\itshape no call --- }no component identified.
\end{taskcardnc}
 
\begin{taskcardnc}{\textbf{ERR\_008}}
\textit{``Add a subcomponent named cpu0 into acme:accel:fft:3.0.''}
\tcblower
\textnormal{\itshape no call --- }child\_component\_id (what to instantiate) is missing.
\end{taskcardnc}
 
\FloatBarrier
\subsection{Cross}\label{app:ds_cross}
\begin{taskcard}{\textbf{CROSS\_001}}
\textit{``Create four CPU shells (vendor Comp, library egg, version 1.0, names cpu1..cpu4).''}
\tcblower
create\_component(Comp, egg, cpu1, 1.0) \\ create\_component(Comp, egg, cpu2, 1.0) \\ create\_component(Comp, egg, cpu3, 1.0) \\ create\_component(Comp, egg, cpu4, 1.0)
\end{taskcard}
 
\begin{taskcard}{\textbf{CROSS\_002}}
\textit{``To all components, add a single input signal port called p1.''}
\\[1pt]\textcolor{black!60}{\small\textit{Note:} "all" = cpu1..cpu4 from CROSS\_001 (one call each).}
\tcblower
add\_signal\_port(p1, in, w=1) \\ add\_signal\_port(p1, in, w=1) \\ add\_signal\_port(p1, in, w=1) \\ add\_signal\_port(p1, in, w=1)
\end{taskcard}
 
\begin{taskcard}{\textbf{CROSS\_003}}
\textit{``In cpu4 add transactional initiator ports ibus1..ibus4.''}
\tcblower
\textnormal{[on Comp:egg:cpu4:1.0]} \\ add\_transactional\_port(ibus1, initiator) \\ add\_transactional\_port(ibus2, initiator) \\ add\_transactional\_port(ibus3, initiator) \\ add\_transactional\_port(ibus4, initiator)
\end{taskcard}
 
\begin{taskcard}{\textbf{CROSS\_004}}
\textit{``To cpu1, add subcomponents: cpu2 as sub1, cpu3 as 
sub2, cpu4 as sub3.''}
\tcblower
\textnormal{[on Comp:egg:cpu1:1.0]} \\ add\_subcomponent(cpu2 as sub1) \\ add\_subcomponent(cpu3 as sub2) \\ add\_subcomponent(cpu4 as sub3)
\end{taskcard}
 
\begin{taskcard}{\textbf{CROSS\_005}}
\textit{``Add to it also the same initiator ports of cpu4.''}
\\[1pt]\textcolor{black!60}{\small\textit{Note:} "it"=cpu1;
replicate cpu4's ibus1..ibus4 (from CROSS\_003).}
\tcblower
\textnormal{[on Comp:egg:cpu1:1.0]} \\ add\_transactional\_port(ibus1, initiator) \\ add\_transactional\_port(ibus2, initiator) \\ add\_transactional\_port(ibus3, initiator) \\ add\_transactional\_port(ibus4, initiator)
\end{taskcard}
 
\begin{taskcard}{\textbf{CROSS\_006}}
\textit{``Hook all ibus ports in sub3 to the same-named ports of the parent (cpu1).''}
\\[1pt]\textcolor{black!60}{\small\textit{Note:} sub3 is an instance of cpu4 (CROSS\_004), which has ibus1..ibus4.}
\tcblower
\textnormal{[on Comp:egg:cpu1:1.0]} \\ connect(sub3.ibus1 -> hierarchical.ibus1) \\ connect(sub3.ibus2 -> hierarchical.ibus2) \\ connect(sub3.ibus3 -> hierarchical.ibus3) \\ connect(sub3.ibus4 -> hierarchical.ibus4)
\end{taskcard}
 
\subsection{Noisy context}\label{app:ds_noise}
Every two design tasks, one external-server (Git or GDB) task is inserted;
the agent must route to the right server. Routing mistakes surface as EVCR, not ECC.\smallskip
 
\begin{taskcard}{\textbf{EASY\_NOISE\_IPXACT\_001}}
\textit{``Create a component with vendor acme, library core, name top, version 1.0.''}
\tcblower
create\_component(acme, core, top, 1.0)
\end{taskcard}
 
\begin{taskcard}{\textbf{EASY\_NOISE\_GIT\_001}}
\textit{``Show the last 5 commits in the repository at /home/user/.../dummy\_repo.''}
\tcblower
git\_log(repo, max\_count=5)
\end{taskcard}
 
\begin{taskcard}{\textbf{EASY\_NOISE\_IPXACT\_002}}
\textit{``Add an input signal port named clk of width 1 to component acme:core:top:1.0.''}
\tcblower
\textnormal{[on acme:core:top:1.0]} \\ add\_signal\_port(clk, in, w=1)
\end{taskcard}
 
\begin{taskcard}{\textbf{EASY\_NOISE\_GDB\_001}}
\textit{``Start a new debugging session.''}
\tcblower
gdb\_start()
\end{taskcard}
 
\begin{taskcard}{\textbf{HARD\_NOISE\_IPXACT\_001}}
\textit{``(same as HARD\_001: create riscv\_tile, add m\_axi and clk, save).''}
\tcblower
create\_component(acme, ip\_cores, riscv\_tile, 2.1.0) \\ add\_transactional\_port(m\_axi, initiator) \\ add\_signal\_port(clk, in, w=1) \\ save\_component
\end{taskcard}
 
\begin{taskcard}{\textbf{HARD\_NOISE\_GIT\_001}}
\textit{``List all remote branches in the repository at 
/home/user/.../dummy\_repo.''}
\tcblower
git\_branch(repo, remote)
\end{taskcard}
 
\begin{taskcard}{\textbf{HARD\_NOISE\_GDB\_001}}
\textit{``Show the source at the current execution point in debugging session 1.''}
\tcblower
gdb\_list\_source(session=1)
\end{taskcard}

\onecolumn
\section{Stability Experiments}
\label{app:stability}

\subsection{Temperature Sweep}
\label{app:temp_results}

\begin{table}[h!]
\centering
\small
\setlength{\tabcolsep}{4pt}
\begin{tabular}{lccccccccc}
& \multicolumn{3}{c}{\textbf{$T=0$}} & \multicolumn{3}{c}{\textbf{$T=0.5$}} & \multicolumn{3}{c}{\textbf{$T{=}\text{default}$}} \\
\cmidrule(lr){2-4}\cmidrule(lr){5-7}\cmidrule(lr){8-10}
\textbf{Model} & ECC & EVCR & TFR & ECC & EVCR & TFR & ECC & EVCR & TFR \\
\midrule
Gemma~4~26B  & 0.896 & 0.055 & 0.094 & 0.889 & 0.055 & 0.111 & \textbf{0.881} & 0.056 & \textbf{0.105} \\
Gemma~4~E4B  & 0.539 & 0.044 & 0.129 & 0.523 & 0.042 & 0.127 & \textbf{0.553} & 0.046 & 0.144 \\
Llama~3.1~8B & \textbf{0.247} & 0.058 & 0.268 & 0.235 & 0.052 & \textbf{0.264} & 0.243 & 0.056 & 0.268 \\
\end{tabular}
\caption{ECC and TFR across temperatures 0, 0.5, and each model's default (1.0, except 0.8 for Llama~3.1~8B), averaged over
the three datasets covered by the sweep (Cross, Errors, Hard) and over
all 16 system-prompt $\times$ tool-description $\times$ history-scope
configurations per model. Rankings and absolute performance are stable
across temperatures for every model; lowering temperature does not
consistently reduce TFR.}
\label{tab:temp_sweep}
\end{table}

\subsection{Run-to-Run Variance}
\label{app:variance_results}
\label{app:stability_tables}

\paragraph{All configurations, hardest suites (2 runs each).}
\begin{table}[h!]
\centering
\small
\setlength{\tabcolsep}{4pt}
\begin{tabular}{lcc}
\textbf{Model} & $\bar\sigma_{\mathrm{ECC}}$ & $\sigma_{\max,\mathrm{ECC}}$ \\
\midrule
Gemma~4~26B  & \textbf{0.025} & 0.076 \\
Llama~3.1~8B & 0.051 & 0.219 \\
Gemma~4~E4B  & 0.059 & \textbf{0.494} \\
\end{tabular}
\caption{Mean and maximum standard deviation of ECC across two repeated
runs of every system-prompt $\times$ tool-description $\times$
history-scope configuration (16 per model), on Cross, Errors, and Hard
--- the three most demanding suites. The Gemma~4~E4B outlier
($\sigma_{\max}=0.494$) is the \texttt{run}-scope, few-shot,
minimal-tools configuration, the same one responsible for the
few-shot collapse reported in Section~\ref{sec:config_effects}, confirming that
effect is genuine rather than a sampling artefact.}
\label{tab:run_variance_full}
\end{table}

\paragraph{Best/worst configuration, repeated subset (10 runs each).}
\begin{table}[h!]
\centering
\small
\setlength{\tabcolsep}{4pt}
\begin{tabular}{llccc}
\textbf{Model} & \textbf{Config} & ECC & EVCR & TFR \\
\midrule
Gemma~4~26B & best  & \textbf{0.973} & 0.027 & \textbf{0.013} \\
Gemma~4~26B & worst & 0.829 & \textbf{0.017} & 0.067 \\
Llama~3.1~8B & best  & \textbf{0.495} & 0.106 & 0.291 \\
Llama~3.1~8B & worst & 0.034 & \textbf{0.001} & \textbf{0.028} \\
\end{tabular}
\caption{Mean ECC, EVCR, and TFR across 10 repeated runs of each
model's best and worst configuration, averaged over all six core
suites (10 tasks each). Standard deviations of ECC across runs: 0.008
(Gemma best), 0.038 (Gemma worst), 0.033 (Llama best), 0.109 (Llama
worst) --- the latter driven by near-total collapse (9 of 10 runs at
ECC$\,\approx\,$0) rather than ordinary run-to-run noise.}
\label{tab:run_variance_subset}
\end{table}

\section{Additional Configuration Results}
\label{app:additional_results}

\subsection{Prompt Ablations}
\label{app:prompt_results}

\begin{table}[H]
\centering
\small
\setlength{\tabcolsep}{4pt}
\resizebox{\textwidth}{!}{%
\begin{tabular}{l*{12}{c}}
& \multicolumn{3}{c}{\textbf{none}} & \multicolumn{3}{c}{\textbf{basic}} & \multicolumn{3}{c}{\textbf{MD}} & \multicolumn{3}{c}{\textbf{few-shot}} \\
\cmidrule(lr){2-4}\cmidrule(lr){5-7}\cmidrule(lr){8-10}\cmidrule(lr){11-13}
\textbf{Model} & ECC & EVCR & TFR & ECC & EVCR & TFR & ECC & EVCR & TFR & ECC & EVCR & TFR \\
\midrule
Gemma~4~31B  & 0.935 & 0.039 & 0.076 & 0.913 & 0.010 & 0.056 & \textbf{0.956} & 0.021 & 0.061 & 0.571 & 0.006 & 0.032 \\
Gemma~4~26B  & 0.907 & 0.044 & 0.086 & 0.910 & 0.020 & 0.063 & \textbf{0.938} & 0.027 & 0.063 & 0.897 & 0.034 & 0.071 \\
GPT-OSS~20B  & \textbf{0.861} & 0.078 
& 0.104 & 0.784 & 0.030 & 0.068 & 0.859 & 0.042 & 0.061 & 0.741 & 0.030 & 0.072 \\
Qwen~3.5~27B & 0.847 & 0.063 & 0.068 & 0.827 & 0.031 & 0.058 & \textbf{0.852} & 0.034 & 0.057 & 0.825 & 0.032 & 0.054 \\
Qwen~3.6~27B & 0.839 & 0.063 & 0.106 & 0.798 & 0.030 & 0.063 & \textbf{0.853} & 0.058 & 0.060 & 0.801 & 0.036 & 0.056 \\
Gemma~4~E4B  & 0.675 & 0.036 & 0.148 & 0.694 & 0.031 & 0.124 & \textbf{0.731} & 0.026 & 0.135 & 0.179 & 0.016 & 0.072 \\
Llama~3.1~8B & {0.293} & 
{0.031} & {0.301} & {0.284} & {0.049} & {0.255} & {0.290} & {0.056} & {0.229} & \textbf{0.339} & {0.034} & {0.211} \\
\end{tabular}%
}
\caption{System-prompt ablation, averaged over the six core suites,
tool-description formats, and history scopes.
The few-shot collapse for
Gemma~4~31B ($0.956\!\to\!0.571$) and Gemma~4~E4B ($0.731\!\to\!0.179$)
is visible as a coverage drop with EVCR and TFR falling together
(inaction, not error).}
\label{tab:prompt}
\end{table}

\subsection{Tool-Description Ablations}
\label{app:tool_description_results}
\begin{table}[H]
\centering
\small
\setlength{\tabcolsep}{4pt}
\begin{tabular}{lcccccc}
& \multicolumn{3}{c}{\textbf{comprehensive}} & \multicolumn{3}{c}{\textbf{minimal}} \\
\cmidrule(lr){2-4}\cmidrule(lr){5-7}
\textbf{Model} & ECC & EVCR & TFR & ECC & EVCR & TFR \\
\midrule
Gemma~4~31B  & 0.819 & 0.019 & 0.031 & \textbf{0.868} & 0.020 & 0.081 \\
Gemma~4~26B  & \textbf{0.932} & 0.028 & 0.048 & 0.894 & 0.034 & 0.093 \\
GPT-OSS~20B  & \textbf{0.859} & 0.046 & 0.057 & 0.764 & 0.044 & 0.096 \\
Qwen~3.5~27B & \textbf{0.845} & 0.034 & 0.040 & 0.830 & 0.046 & 0.078 \\
Qwen~3.6~27B & \textbf{0.823} & 0.037 & 0.046 
& 0.822 & 0.057 & 0.097 \\
Gemma~4~E4B  & \textbf{0.610} & 0.028 & 0.073 & 0.530 & 0.026 & 0.167 \\
Llama~3.1~8B & \textbf{0.312} & {0.050} & {0.225} & {0.290} & {0.035} & {0.273} \\
\end{tabular}
\caption{Tool-description ablation, averaged over the six core suites,
system prompts, and history scopes.
Minimal descriptions raise TFR for
every model (roughly doubling it in most cases);
the ECC effect is
less uniform.}
\label{tab:tools_effect}
\end{table}

\subsection{History-Scope Ablations}
\label{app:history_results}
\begin{table}[H]
\centering
\small
\setlength{\tabcolsep}{4pt}
\begin{tabular}{lcccccc}
& \multicolumn{3}{c}{\textbf{run}} & \multicolumn{3}{c}{\textbf{task}} \\
\cmidrule(lr){2-4}\cmidrule(lr){5-7}
\textbf{Model} & ECC & EVCR & TFR & ECC & EVCR & TFR \\
\midrule
Gemma~4~31B  & \textbf{0.858} & 0.014 & 0.037 & 0.829 & 0.024 & 0.075 \\
Gemma~4~26B  & \textbf{0.918} & 0.029 & 0.071 & 0.909 & 0.034 & 0.071 \\
GPT-OSS~20B  & \textbf{0.862} & 0.051 & 0.061 & 0.760 & 0.039 & 0.091 \\
Qwen~3.5~27B & \textbf{0.864} & 0.047 & 0.038 & 0.811 & 0.033 & 0.081 \\
Qwen~3.6~27B & \textbf{0.823} & 0.051 & 0.060 & 0.822 & 0.042 & 0.083 \\
Gemma~4~E4B  & \textbf{0.621} & 0.031 & 0.125 & 0.519 
& 0.023 & 0.115 \\
Llama~3.1~8B & {0.157} & {0.015} & {0.084} & \textbf{0.445} & 0.070 & 0.413 \\
\end{tabular}
\caption{History-scope ablation, averaged over the six core suites,
system prompts, and tool-description formats.
Llama~3.1~8B is the only
model that prefers per-task scope; all others tolerate or benefit from
cumulative history at this session length.}
\label{tab:scope}
\end{table}

\section{Tool Description Example}
\label{app:tool_example}
The following shows the \texttt{add\_signal\_port} tool in both description formats.

\begin{promptbox}{\texttt{add\_signal\_port} --- comprehensive}
Add a signal port to an existing component.

    The component must already exist and the port name must be unique.
    Direction must be 'in', 'out', or 'inout'.
    Width and multiplicity must be greater than zero. Default multiplicity value is 1.

    On success, the new signal port becomes available for future connections.
\end{promptbox}

\begin{promptbox}{\texttt{add\_signal\_port} --- minimal}
Add a signal port to a component.
\end{promptbox}

\twocolumn
\section{System Prompts}
\label{app:prompts}

All prompts are reproduced verbatim.
Section~\ref{sec:react-prompts} contains the ReAct system prompts
(\texttt{none}, \texttt{basic}, \texttt{MD}, \texttt{fewshot}).
Section~\ref{sec:planact-prompts} contains the Plan-and-Act stage
prompts: planner (\texttt{basic}, \texttt{structured}), validator
(\texttt{basic}, \texttt{structured}), and worker (fixed).
\subsection{ReAct Prompts}
\label{sec:react-prompts}

\begin{promptbox}{\texttt{none}}
\textnormal{No system prompt. The model receives only the task instruction and the tool schemas.}
\end{promptbox}

\begin{promptbox}{\texttt{basic}}
You are an autonomous system-design agent operating in a tool-enabled environment.
The server exposes tools that create, inspect, connect, delete, load, and persist components and their internal elements.
\medskip

If the user requests creation, modification, connection, deletion, loading, or persistence of components or elements,
you MUST call the appropriate tool when all required arguments are available.
\medskip

If the user asks to rename or edit an element and there is no direct update tool for that operation,
treat the request as deleting the existing element and recreating the same element with the requested change,
while keeping all other retrievable fields unchanged.
For example, changing the multiplicity of a signal port means deleting that port and recreating it with the new multiplicity and the same remaining attributes.
\medskip

If the request appears incomplete, then follow this rule:
if you are confident that the request contains a conceptual error, do NOT call any tool and explain the problem;
if you are unsure, first use inspection tools (e.g., list\_components, get\_component\_details, or get\_element) and then decide how to proceed.
\medskip

If some required tool arguments are missing after relevant inspection,
DO NOT invent values and DO NOT call the tool.
Instead, inform the user which required fields are missing.
\medskip

You MUST NOT respond in natural language when a tool can be correctly used.
\end{promptbox}

\begin{promptbox}{\texttt{MD}}
\textbf{\#\# Role}\\
You are an autonomous design agent operating in a tool-enabled environment.
The server exposes tools that create, inspect, connect, delete, load, and persist components and their elements.
\medskip

\textbf{\#\# Behavior}\\
- Call one tool per turn.\\
- Do not invent argument values.
Use only values stated in the task or retrieved from tools.\\
- If required arguments are missing after inspection, do not call the tool.\\
- Complete all operations stated in the task --- do not stop after the first tool call.\\
- If the request is invalid or impossible, say so clearly without calling any tool.
\medskip

\textbf{\#\# Hierarchical connections}\\
Use "hierarchical" as the instance name only when the endpoint is a top-level port
of the parent component itself, not a port of a subcomponent.
\medskip

\textbf{\#\# Invalid tool calls}\\
If a tool call fails, do not retry with the same arguments.
Read the error message and adjust before trying again.
\end{promptbox}

\begin{promptbox}{\texttt{fewshot}}
You are an autonomous system-design agent operating in a tool-enabled environment.
The server exposes tools that create, inspect, connect, delete, load, and persist components and their elements.
If the user requests creation, modification, connection, deletion, loading, or persistence of components or elements,
you MUST call the appropriate tool when all required arguments are available.
If required arguments are missing after inspection, do NOT invent values --- inform the user instead.
You MUST NOT respond in natural language when a tool can be correctly used.
\medskip

---
\medskip

EXAMPLE 1 --- Apply the same operation to every component listed. Do not stop early.
\medskip

User: Add a 1-bit input signal port named clk to each of these components: nordic:rf:tx\_core:1.0, nordic:rf:rx\_core:1.0, nordic:rf:pll:1.0.
\medskip

{[}calls add\_signal\_port: component\_id=nordic:rf:tx\_core:1.0,\allowbreak port\_name=clk,\allowbreak direction=in,\allowbreak width=1,\allowbreak multiplicity=1{]}\\
Tool: \{"ok": true\}

{[}calls add\_signal\_port: component\_id=nordic:rf:rx\_core:1.0,\allowbreak port\_name=clk,\allowbreak direction=in,\allowbreak width=1,\allowbreak multiplicity=1{]}\\
Tool: \{"ok": true\}

{[}calls add\_signal\_port: component\_id=nordic:rf:pll:1.0,\allowbreak port\_name=clk,\allowbreak direction=in,\allowbreak width=1,\allowbreak multiplicity=1{]}\\
Tool: \{"ok": true\}

A: Done.
Port clk added to all three components.
\medskip

---
\medskip

EXAMPLE 2 --- When a task has multiple distinct operations, complete all of them.
\medskip

User: Inside arm:soc:top:2.0, connect cpu\_inst's axi\_out port to bus\_inst's axi\_in. Also connect the top-level sys\_rst signal to cpu\_inst's rst\_n.
Then save.
\medskip

{[}calls add\_subcomponent\_connection: parent\_component\_id=arm:soc:top:2.0,\allowbreak from\_instance=cpu\_inst,\allowbreak from\_port=axi\_out,\allowbreak to\_instance=bus\_inst,\allowbreak to\_port=axi\_in{]}\\
Tool: \{"ok": true\}

{[}calls add\_subcomponent\_connection: parent\_component\_id=arm:soc:top:2.0,\allowbreak from\_instance=hierarchical,\allowbreak from\_port=sys\_rst,\allowbreak to\_instance=cpu\_inst,\allowbreak to\_port=rst\_n{]}\\
Tool: \{"ok": true\}

{[}calls save\_component: component\_id=arm:soc:top:2.0{]}\\
Tool: \{"ok": true\}

A: Both connections established and component saved.
\medskip

---
\medskip

\textbf{EXAMPLE 3 --- The \texttt{"hierarchical"} instance name.}
\medskip

\texttt{"hierarchical"} refers exclusively to a port that belongs to the parent component itself, not to any subcomponent.
Use it when wiring a top-level port down to a subcomponent, or up from a subcomponent to the top level.
Never use it to connect two subcomponent instances to each other.
\medskip

User: In risc:sys:chip\_top:1.0, wire the top-level ext\_irq input down to irq\_ctrl\_inst's irq\_in port.
Then wire irq\_ctrl\_inst's status\_out up to the top-level debug\_out port.
\medskip

{[}calls add\_subcomponent\_connection: parent\_component\_id=risc:sys:chip\_top:1.0,\allowbreak from\_instance=hierarchical,\allowbreak from\_port=ext\_irq,\allowbreak to\_instance=irq\_ctrl\_inst,\allowbreak to\_port=irq\_in{]}\\
Tool: \{"ok": true\}

{[}calls add\_subcomponent\_connection: parent\_component\_id=risc:sys:chip\_top:1.0,\allowbreak from\_instance=irq\_ctrl\_inst,\allowbreak from\_port=status\_out,\allowbreak to\_instance=hierarchical,\allowbreak to\_port=debug\_out{]}\\
Tool: \{"ok": true\}

A: Both hierarchical connections established.
\medskip

---
\medskip

EXAMPLE 4 --- If add\_subcomponent\_connection fails, do NOT modify child components or retry with guessed port names.
Check get\_component\_details on the parent to verify what subcomponents and connections exist, then report what is missing.
Do not invent ports or restructure components to force a connection.
\medskip

User: In mips:platform:board:3.0, connect dma\_inst's mem\_req port to ram\_inst's mem\_ack port.
\medskip

{[}calls add\_subcomponent\_connection: parent\_component\_id=mips:platform:board:3.0,\allowbreak from\_instance=dma\_inst,\allowbreak from\_port=mem\_req,\allowbreak to\_instance=ram\_inst,\allowbreak to\_port=mem\_ack{]}\\
Tool: \{"ok": false, "error": \{"code": "invalid\_source\_port", "message": "Port mem\_req not found on instance dma\_inst."\}\}

{[}calls get\_component\_details: component\_id=mips:platform:board:3.0{]}\\
Tool: \{"ok": true, "subcomponents": ["dma\_inst", "ram\_inst"], "connections": []\}

A: The connection failed because port mem\_req does not exist on dma\_inst.
I cannot create the connection. Please verify the correct port names on dma\_inst and ram\_inst.
\medskip

---
\medskip

Apply these patterns to every task. Always apply an operation to every component or element the task specifies --- do not stop early.
Always complete every distinct operation in the task before finishing.
Always call save\_component after structural changes when the task requests persistence.
Use "hierarchical" only for ports that belong to the parent component itself.
If a connection fails, inspect the parent with get\_component\_details and report what is missing --- do not modify child components or invent port names.
\end{promptbox}

\subsection{Plan-and-Act Prompts}
\label{sec:planact-prompts}

\begin{promptbox}{Planner \texttt{basic}}
You are a planning agent responsible for decomposing a user request into a sequence of executable steps.
Each step will be executed by a separate worker agent that has access to the following tools:
\{tools\}
\medskip

RULES:\\
- Each step must correspond to exactly ONE tool call.\\
- Steps must be ordered so that any value produced by step N is available to step N+1.\\
- If a step produces an identifier, name, or value that a later step needs, say so explicitly in that later step (e.g. `using the component ID returned in the previous step').\\
- Preserve all identifiers, names, and parameters from the user request exactly as given.\\
- Do not invent missing values.\\
- Avoid pronouns like `it' or `that' --- always 
repeat the full name or identifier.\\
- If the task requires only one tool call, output one step.
\medskip

OUTPUT FORMAT:\\
- Write steps in natural language.\\
- Separate steps with the token {\textless}STEP{\textgreater} placed at the END of each step, including the last one.\\
- Do not number the steps.\\
- Do not add any preamble, explanation, or summary --- output only the steps.
\medskip

Example for a two-step task:\\
Create a component with id riscv\_core, vendor openhw, library cpu, and version 1.0.{\textless}STEP{\textgreater}\\
Add a port named data\_bus to the component with id riscv\_core, using protocol transactional and port type initiator.{\textless}STEP{\textgreater}
\end{promptbox}

\begin{promptbox}{Planner \texttt{structured}}
You are a planning agent.
Your job is to decompose a user request into a sequence of atomic steps, each executed by a separate worker agent.
The worker has access to these tools --- use them to understand what operations are possible, but write your steps in plain natural language describing the action, not the tool call:
\{tools\}
\medskip

STEP RULES:\\
- Each step must correspond to exactly ONE tool call.\\
- If a task needs 4 tool calls, write 4 steps.
Never collapse multiple tool calls into one step.\\
- Order steps so that any value produced by step N is available to step N+1.\\
- When a later step needs an ID or value from an earlier step, say so explicitly (e.g. `using the component ID returned in step 1').\\
- Copy all identifiers, names, and parameters from the user request exactly.\\
- Do not invent missing values.\\
- Never use pronouns like `it' or `that' --- always repeat the full name.
\medskip

RENAME AND EDIT OPERATIONS:\\
- There is no rename or update tool.
Any request to rename or change a field of an existing element must be planned as: retrieve, delete, recreate.\\
- If the current field values are not given in the task, add a step to retrieve the element with get\_element before the delete step.\\
- The recreate step must preserve all fields unchanged except the one being modified.\\
- Example: renaming signal port p2 to p\_out on component X:\\
\quad 1. Retrieve port p2 from component X to get its current fields.{\textless}STEP{\textgreater}\\
\quad 2. Delete port p2 from component X.{\textless}STEP{\textgreater}\\
\quad 3. Add a signal port named p\_out to component X with the same direction, width, 
and multiplicity as retrieved in step 1.{\textless}STEP{\textgreater}
\medskip

INSPECT-BEFORE-CONNECT:\\
- If the instance names or port names needed for a connection step are not already known from the task description or prior steps, include a prior step that retrieves the component details first.\\
- Never guess or assume instance or port names.
\medskip

THE "hierarchical" INSTANCE NAME --- READ CAREFULLY:\\
- "hierarchical" is a reserved instance name meaning the parent component itself, not any subcomponent instance.\\
- Use it ONLY when one endpoint of a connection is a port that belongs directly to the parent component.\\
- CORRECT: connecting subcomponent sub3 port ibus1 outward to the parent's own port ibus1:\\
\quad from\_instance="sub3",\allowbreak from\_port="ibus1",\allowbreak to\_instance="hierarchical",\allowbreak to\_port="ibus1"\\
- WRONG: connecting subcomponent bus\_inst to subcomponent ram\_inst:\\
\quad WRONG: from\_instance="hierarchical",\allowbreak from\_port="m0\_port",\allowbreak to\_instance="ram\_inst",\allowbreak to\_port="s\_axi"\\
\quad RIGHT: from\_instance="bus\_inst",\allowbreak from\_port="m0\_port",\allowbreak to\_instance="ram\_inst",\allowbreak to\_port="s\_axi"
\medskip

PURE-INSPECTION OR ERROR TASKS:\\
- If the task only requires reading or reporting state with no mutations, output zero steps.\\
- If the task references elements that do 
not exist, output zero steps.\\
- A plan with zero {\textless}STEP{\textgreater} tokens is valid.
\medskip

OUTPUT FORMAT:\\
- Write steps in plain natural language.\\
- End every step, including the last, with the token {\textless}STEP{\textgreater}.\\
- Do not number steps.\\
- Do not add any preamble, heading, explanation, or summary.\\
- If the plan is empty, output nothing at all.
\end{promptbox}

\begin{promptbox}{Validator \texttt{basic}}
You are validating a plan for the following task:
\{task\_prompt\}

Available tools: \{tool\_names\}

Plan to validate:
\{plan\_text\}
\medskip

Check the plan against these criteria:\\
1. Every step maps to at least one available tool.\\
2. No step invents arguments not present in the task or derivable from prior steps.\\
3. Steps are in a logical order (no step depends on a result not yet produced).\\
4. No steps are duplicated or contradictory.
\medskip

If the plan passes all criteria, respond with exactly: VALID\\
If it fails any criterion, respond with: INVALID: {\textless}brief reason{\textgreater}
\end{promptbox}

\begin{promptbox}{Validator \texttt{structured}}
You are validating a plan for the following task:
\{task\_prompt\}

Available operations (the worker uses these tools):
\{tool\_names\}

Plan to validate:
\{plan\_text\}
\medskip

VALIDATION RULES:
\medskip

1. TOOL COVERAGE --- each step must describe an action that is achievable with at least one of the available tools.
Steps are written in natural language: do not require them to contain the exact tool name.
\medskip

2. ARGUMENT VALIDITY --- no step may invent an argument value that is neither given in the task nor derivable from the result of a prior step.
Cross-step references such as `using the component ID returned in the previous step' are explicitly allowed and correct.
\medskip

3. ORDERING --- steps must be in a logical sequence where prerequisites come before the operations that depend on them.
Do NOT reject a step merely because it references a value produced by an earlier step.
\medskip

4. NO INVENTED REQUIREMENTS --- only reject the plan for what it does wrong or omits relative to the task.
Do not add requirements that are not in the task.
\medskip

5. EMPTY PLAN --- a plan with no steps is valid when the task requires no mutations.
Do not reject an empty plan solely because it has no steps.
\medskip

Respond with exactly one of:\\
\quad VALID\\
\quad INVALID: {\textless}one concise reason referencing the specific step or argument that fails{\textgreater}
\end{promptbox}

\begin{promptbox}{Worker prompt (fixed, not an experimental axis)}
You are an execution agent operating in a tool-enabled environment.
You are executing exactly ONE step of a larger plan.
\medskip

Your context contains:\\
- Results from previous steps (if any) --- use those values directly, do not re-derive them.\\
- The current step you must execute.
\medskip

SCOPE RULES:\\
- Execute ONLY what the current step explicitly describes.
Nothing more.\\
- Do NOT inspect components, add ports, modify state, or call any tool that the step does not explicitly request.\\
- Do NOT attempt to fix or improve results from previous steps.\\
- If the step says to retrieve component details, call that tool only and report what it returned.
Do not act further on the result.\\
- Call at most one tool per step unless the step explicitly lists two actions.
\medskip

ARGUMENT RULES:\\
- Use only argument values stated in the step or present in prior step results.\\
- Do not invent, guess, or supply default values for missing arguments.\\
- If a required argument is missing, do NOT call the tool.
Report which argument is missing.
\medskip

RENAME AND EDIT OPERATIONS:\\
- There is no rename or update tool.
If the step says to rename or edit an element, execute it as delete followed by recreate.\\
- Use only the field values present in the step or prior step results --- never drop unaffected fields when recreating.
\medskip

THE "hierarchical" INSTANCE NAME:\\
- "hierarchical" means the parent component itself, not a subcomponent.\\
- Use it only when the step explicitly says the endpoint is a top-level port of the parent component.\\
- If a step connects two subcomponent instances to each other, use their real instance names --- never use "hierarchical" for a subcomponent endpoint.
\medskip

ON FAILURE:\\
- If the tool returns an error, do NOT retry with the same or arbitrarily varied arguments.\\
- Only retry if the error message gives clear actionable information about what to correct.\\
- If the error does not give enough information to fix the call, report failure immediately.
\medskip

After the tool call completes, write exactly one line:\\
\quad Step result: {\textless}what was done and the key output value, e.g.
an ID or status{\textgreater}

On failure, write exactly:\\
\quad Step result: FAILED --- {\textless}reason from the tool response{\textgreater}

Do not write anything beyond that line.
\end{promptbox}

\end{document}